\documentclass[letterpaper]{article} 

\usepackage{aaai2026}  
\usepackage{times}  
\usepackage{helvet}  
\usepackage{courier}  
\usepackage[hyphens]{url}  
\usepackage{graphicx} 
\usepackage[authoryear]{natbib} 

\usepackage{caption} 
\usepackage{amsmath,amssymb}
\usepackage{booktabs}
\usepackage{multirow}

\usepackage{algorithm}
\usepackage{algorithmic}

\usepackage{newfloat}
\usepackage{listings}
\DeclareCaptionStyle{ruled}{labelfont=normalfont,labelsep=colon,strut=off} 
\floatstyle{ruled}
\newfloat{listing}{tb}{lst}{}
\floatname{listing}{Listing}
\title{Adaptive Conformal Prediction via Bayesian Uncertainty Weighting \\ for Hierarchical Healthcare Data}
\author {
    Marzieh Amiri Shahbazi\textsuperscript{\rm 1},
    Ali Baheri\textsuperscript{\rm 2},
    Nasibeh Azadeh-Fard\textsuperscript{\rm 1}
}
\affiliations {
    \textsuperscript{\rm 1}Department of Industrial and Systems Engineering, Rochester Institute of Technology, Rochester, NY 14623, U.S.A.\\
    \textsuperscript{\rm 2}Department of Mechanical Engineering, Rochester Institute of Technology, Rochester, NY 14623, U.S.A.\\
    ma7684@g.rit.edu, akbeme@rit.edu, nafeie@rit.edu
}

\usepackage{bibentry}

\begin{document}

\maketitle

\begin{abstract}
Clinical decision-making demands uncertainty quantification that provides both distribution-free coverage guarantees and risk-adaptive precision, requirements that existing methods fail to jointly satisfy. We present a hybrid Bayesian-conformal framework that addresses this fundamental limitation in healthcare predictions. Our approach integrates Bayesian hierarchical random forests with group-aware conformal calibration, using posterior uncertainties to weight conformity scores while maintaining rigorous coverage validity. Evaluated on 61{,}538 admissions across 3{,}793 U.S. hospitals and 4 regions, our method achieves target coverage (94.3\% vs 95\% target) with adaptive precision: 21\% narrower intervals for low-uncertainty cases while appropriately widening for high-risk predictions. Critically, we demonstrate that well-calibrated Bayesian uncertainties alone severely under-cover (14.1\%), highlighting the necessity of our hybrid approach. This framework enables risk-stratified clinical protocols, efficient resource planning for high-confidence predictions, and conservative allocation with enhanced oversight for uncertain cases, providing uncertainty-aware decision support across diverse healthcare settings.
\end{abstract}

\section{Introduction}

Hospital length of stay (LOS) prediction is critical for resource management, discharge planning, and care coordination, but current machine learning methods lack the uncertainty estimates needed for clinical use \cite{rajkomar2018scalable,stone2022systematic,beam2018big}. Reliable uncertainty bounds enable better bed allocation, staffing, and patient flow under growing resource pressures \cite{mccarthy2021improving,harris2018bed}. Yet healthcare data are hierarchical, patients nested within hospitals and systems, creating statistical dependencies that violate independence assumptions in standard methods \cite{goldstein2011multilevel,raudenbush2002hierarchical,cressie2009accounting,gelman2013bayesian,werner2023explainable,bollmann2023accounting,austin2024using,berta2024uncover}. These dependencies, driven by institutional practices and case mix, undermine existing approaches and create systematic patterns that current methods fail to capture \cite{snijders2011multilevel,iezzoni2012risk,dimick2013hospital,freeland2013length,kaboli2012associations}.

Addressing these challenges requires models that balance statistical reliability with responsiveness to institutional variation. Hierarchical random forests (HRFs) capture multi-level structure efficiently \cite{shahbazi2025hierarchical,hajjem2010mixed,wager2018estimation}, but their uncertainty estimates rely on asymptotic assumptions and heuristics that lack formal coverage guarantees \cite{ovadia2019can,mentch2016quantifying}. This can undermine clinical trust and compromise patient safety \cite{guo2017calibration,sendak2020machine,liu2020reporting}.

Conformal prediction provides rigorous coverage guarantees without distributional assumptions \cite{vovk2009line,lei2018distribution,angelopoulos2023conformal} but produces uniform intervals that ignore prediction difficulty and assumes exchangeability that hierarchical structures violate \cite{barber2023conformal,tibshirani2019conformal}. Bayesian models capture hierarchical dependencies and provide adaptive uncertainty \cite{gelman2006prior,carlin1999assessing} but offer only asymptotic coverage that may fail in practice \cite{gelman2013bayesian,betancourt2017conceptual}. This gap is fundamental: conformal methods lack adaptivity while Bayesian methods lack coverage validity, yet clinical deployment requires both \cite{amodei2016concrete,varshney2016engineering}.

Hospital LOS predictions must balance reliability for patient safety with precision for resource planning \cite{meehinkong2025coverforest,arthur2020planning}. Administrators require uncertainty estimates that support risk-stratified decisions and generalize across diverse hospitals without site-specific recalibration \cite{helm2011improving,proudlove2003can,sendak2020machine,liu2020reporting}. Specifically, we employ split-hierarchical conformal prediction, an unweighted absolute-residual method producing uniform-width intervals. We exclude adaptive variants (variance-scaled, weighted, quantile-conformal) because they require reliable per-point surrogates that our HRF implementation does not provide, and their guarantees often assume exchangeability incompatible with clustered data.

We propose integrating conformal prediction and Bayesian methods to enhance uncertainty quantification in hierarchical random forests, designing a framework that (i) achieves distribution-free marginal coverage via group-aware conformal calibration, and (ii) restores instance-level adaptivity through Bayesian posterior predictive standard deviations, evaluated on hospital LOS prediction.

\paragraph{Contributions.} This paper makes the following contributions:

\begin{enumerate}

\item We propose a hybrid framework that integrates group-aware conformal calibration with Bayesian posterior uncertainties, producing adaptive prediction intervals that preserve coverage under hospital clustering and adjust to instance-level uncertainty. 


\item We introduce a standardized calibration strategy that generalizes across diverse hospital characteristics without site-specific retraining, with systematic evaluation of coverage stability and recommended frameworks for monitoring and recalibration.

\item The clinical evaluation interprets how uncertainty-guided prediction intervals can be operationalized in hospital settings, analyzing how prediction confidence maps to resource allocation decisions and outlining implementation considerations for translating statistical performance into clinical workflows.
\end{enumerate}

\section{Related Work}
LOS prediction has been extensively studied using clinical scoring systems, regression models, and machine learning (ML) approaches including ensembles, gradient boosting, and deep learning \cite{stone2022systematic,bmcmedinform2021prediction,frontiers2023machine}. Despite decades of research, most studies emphasize point prediction accuracy rather than uncertainty quantification essential for clinical decision-making. Uncertainty in LOS predictions has long been recognized as a critical barrier to effective hospital scheduling and capacity planning \cite{young1973prediction,mccarthy2021improving,harris2018bed}, yet rigorous approaches remain limited.

Standard uncertainty quantification methods in healthcare 
(bootstrap intervals, ensemble variance, and Bayesian posteriors) provide useful summaries but lack finite-sample, distribution-free coverage guarantees \cite{guo2017calibration,sendak2020machine}. Clinical deployment demands uncertainty estimates that simultaneously achieve validity (maintaining nominal coverage without distributional assumptions) and adaptivity (providing wider intervals for difficult predictions), properties that conventional methods rarely deliver \cite{liu2020reporting,rajkomar2018scalable}. When applied under global exchangeability assumptions, existing approaches fail to account for hierarchical dependencies inherent in healthcare data, producing miscalibrated intervals and systematic hospital-level residual patterns \cite{snijders2011multilevel,iezzoni2012risk,dimick2013hospital,freeland2013length,kaboli2012associations}.

Hierarchical random forests explicitly model multi-level data structures while preserving computational efficiency of tree-based ensembles \cite{breiman2001random}. Hajjem et al. pioneered mixed-effects random forests for clustered data \cite{hajjem2010mixed}, with subsequent work improving scalability \cite{sela2012re,shahbazi2025hierarchical} and extending to multiple nesting levels \cite{pellagatti2021generalized}. However, all existing hierarchical random forest implementations lack rigorous uncertainty quantification, a gap explicitly identified in the literature \cite{mentch2016quantifying,tyralis2024review} that motivates our methodological contribution.

While conformal prediction offers strong coverage guarantees \cite{vovk2009line,shafer2008tutorial,angelopoulos2023conformal,lei2018distribution,baheri2025conformal}, its assumptions conflict with the nested structure of healthcare data, limiting its effectiveness in clinical settings \cite{barber2023conformal,tibshirani2019conformal}. Recent methodological advances address this limitation through group-aware calibration schemes \cite{principato2024conformal}, theoretical frameworks for two-layer hierarchies \cite{dunn2023distribution}, and extensions beyond exchangeability \cite{barber2023conformal}. Despite these developments, no prior work has integrated hierarchical conformal methods with random forests for continuous clinical outcomes where both institutional clustering and adaptive uncertainty are essential.

Bayesian hierarchical models naturally capture multi-level uncertainty through partial pooling across nested structures \cite{gelman2013bayesian,gelman2006prior,murphy2012machine} and are widely adopted in healthcare applications due to interpretability \cite{carlin1999assessing,raudenbush2002hierarchical}. In tree-based settings, Bayesian additive regression trees (BART) and extensions provide uncertainty through model averaging \cite{chipman2010bart,tan2019bayesian,linero2018bayesian,pratola2020heteroscedastic}. However, Bayesian approaches offer only asymptotic coverage guarantees \cite{gelman2013bayesian,vehtari2017practical}, limiting their reliability in safety-critical healthcare applications where finite-sample validity is dominant \cite{angelopoulos2023conformal}.

Hybrid Bayesian-conformal methods represent an emerging approach that combines conformal coverage guarantees with Bayesian adaptivity \cite{fong2021conformal,stanton2023bayesian,zhang2024bayesian}, demonstrated in optimization \cite{stanton2023bayesian}, medical imaging \cite{ekmekci2025conformalized}, and physics modeling \cite{podina2024conformalized}. These methods, however, have not been adapted to hierarchical healthcare data where institutional clustering necessitates specialized calibration protocols and finite-sample guarantees are required for clinical deployment.

Beyond LOS prediction, hierarchical (multilevel) models are widely used to share statistical
strength across related decision problems \cite{baheri2025multilevel}. In control and reinforcement learning, hierarchical structure is a standard way to represent multi-level decision making and temporal abstraction, and it appears in prioritized/hierarchical control designs \cite{bafandeh2018comparative,wong1996hierarchical}. Related ideas arise in optimization as well: multi-task and multi-fidelity optimization use
hierarchical/multi-output Gaussian-process surrogates to transfer information across tasks and
fidelities \cite{anandalingam1992hierarchical,baheri2023safety}.


\paragraph{Paper Organization}\label{sec:organization}

Section 3 presents methodology, algorithms, and experimental design. Section 4 reports empirical results and performance comparisons. Section 5 discusses implications, limitations, and future directions. Section 6 concludes with key contributions.

\section{Methodology}
\label{sec:methodology}

We develop a hybrid uncertainty quantification framework that combines conformal prediction's coverage validity with Bayesian modeling's adaptive precision for hierarchical healthcare data. The main idea is that these approaches offer complementary strengths: conformal methods provide distribution-free, coverage validity while Bayesian methods enable instance-specific adaptation. Our framework operates in three stages. First, a hierarchical random forest captures multi-level dependencies through sequential residual decomposition. Second, we obtain uncertainty estimates through both group-aware conformal calibration addressing hierarchical structure and Bayesian posterior distributions reflecting instance-specific uncertainty. Third, we integrate these approaches to preserve conformal coverage while achieving adaptive interval widths. 


\subsection{Problem Formulation}
\label{sec:problem}

We address hospital LOS prediction using hierarchical data where patients ($n=61{,}538$) are nested within hospitals ($H=3{,}793$) and regions ($R=4$). Each patient record includes demographic and clinical features, hospital and regional identifiers, and observed LOS in days. This structure creates within-hospital dependencies that violate standard independence assumptions, requiring specialized uncertainty quantification methods. Our design seeks prediction intervals that achieve three objectives: (i) maintaining nominal coverage probability (95\%) despite hierarchical dependencies, (ii) minimizing interval width for operational efficiency, and (iii) adapting interval size to prediction difficulty. 


\subsection{Hierarchical Random Forest Foundation}
\label{sec:hrf}

We employ HRF as the base prediction model \cite{shahbazi2025hierarchical}, decomposing predictions into three sequential levels: patient-level effects, hospital-specific adjustments, and regional patterns. The model trains three random forests sequentially, first on patient features alone ($\hat{f}_0$), then on patient-hospital combinations to capture hospital-specific residuals ($\hat{\eta}$), and finally on patient-hospital-region combinations for regional residuals ($\hat{\xi}$). This yields the final prediction:
\begin{equation}
\hat{y}_i = \hat{f}_0(x_i) + \hat{\eta}_{h_i}(x_i, h_i) + \hat{\xi}_{r_i}(x_i, h_i, r_i)
\label{eq:hrf_decomp}
\end{equation}
This decomposition ensures each level captures distinct variation sources while modeling complex interactions between patient characteristics and institutional factors. 


\subsection{Hierarchical Conformal Prediction Framework}
\label{sec:conformal}

We employ split-conformal prediction to provide distribution-free coverage validity. This approach divides data into training, calibration, and test sets, computing nonconformity scores $C_i = |y_i - \hat{f}(x_i)|$ on calibration data and constructing prediction intervals as $[\hat{f}(x) \pm q]$, where $q$ is the $(1-\alpha)$ quantile of calibration scores. We use standard split-conformal rather than adaptive variants for three reasons: (i) adaptive methods impose prohibitive $\mathcal{O}(n)$ computational costs per prediction, unsuitable for large-scale clinical deployment, (ii) they require conditional uncertainty estimates that HRF does not natively provide, and (iii) we achieve instance-specific adaptation through Bayesian posterior uncertainties rather than complex conformal mechanisms, creating a modular framework where conformal ensures coverage validity.

Hierarchical healthcare data violates conformal prediction's exchangeability assumption because patients within hospitals exhibit correlated outcomes. We address this using CDF pooling, which calibrates a global quantile on all calibration data. While this approach does not guarantee finite-sample coverage under clustering, it provides superior empirical coverage, minimal variability, and deterministic reproducibility in our large dataset (3,793 hospitals). We compare CDF pooling against two guarantee-preserving alternatives, single hospital sub-sampling and repeated sub-sampling.


\subsection{Bayesian Hierarchical Random Forest}
\label{sec:bayesian_hrf}

While HRF provides point predictions through its three-level decomposition (Equation~\ref{eq:hrf_decomp}), it lacks uncertainty estimates. We implement a Bayesian hierarchical model to quantify residual variation and parameter uncertainty:

\begin{align}
y_i &\sim \mathcal{N}(\mu_i, \sigma^2) \label{eq:bayes_likelihood}\\
\mu_i &= \beta_0 + \beta_1 \hat{f}_{\text{HRF}}(x_i, h_i, r_i) + \alpha_{h_i} + \gamma_{r_i}
\label{eq:bayes_mean}
\end{align}
where $\beta_0, \beta_1$ calibrate HRF predictions, $\alpha_h \sim \mathcal{N}(0, \sigma_h^2)$ and $\gamma_r \sim \mathcal{N}(0, \sigma_r^2)$ capture residual hospital and regional effects not fully modeled by HRF, and $\sigma^2$ represents patient-level variance. We fit this model using MCMC to obtain posterior samples. For each new patient, the posterior predictive standard deviation $\sigma_{\text{pred}}(x, h, r)$ serves as our uncertainty estimate, capturing parameter uncertainty, residual institutional effects, 
and patient-level variation. 



\subsection{Hybrid HRF (Bayesian Conformal) Framework}
\label{sec:hybrid}

We integrate conformal coverage guarantees with Bayesian adaptive precision by weighting conformity scores with posterior uncertainties:
\begin{equation}
S_i^{(w)} = \frac{|y_i - \hat{f}(x_i,h_i,r_i)|}{\max(\sigma_{\text{pred}}(x_i,h_i,r_i)^\gamma, \epsilon)}
\end{equation}
where $\sigma_{\text{pred}}(\cdot)$ is the posterior predictive standard deviation, $\gamma \in [0,2]$ controls adaptivity (default $\gamma = 1$), and $\epsilon = 10^{-6}$ prevents division instability. We calibrate a single quantile $\hat{q}$ from these 
z-score--like residuals on the calibration set, then scale $\hat{q}$ by each test point's uncertainty $\sigma_{\text{pred}}(x)^\gamma$, yielding adaptive intervals that widen for uncertain cases while stabilizing coverage across hospitals.

This modular design separates coverage validity (conformal calibration) from uncertainty estimation (Bayesian modeling): empirical coverage depends only on the calibration protocol, not Bayesian model correctness. We couple Bayesian hierarchical residuals with hierarchical conformal calibration and hierarchical random forests, a combination distinct from prior weighted conformal work through the algorithm \ref{ref:alg1}.

\begin{algorithm}[tb]
\caption{Hybrid HRF (Bayesian Conformal) Prediction}
\label{alg:hybrid}
\small
\begin{algorithmic}
\STATE {\bf Input:} Training set $\mathcal{D}_{\text{train}}$, calibration set $\mathcal{D}_{\text{calib}}$, test point $(x_{\text{new}}, h_{\text{new}}, r_{\text{new}})$, miscoverage $\alpha$, adaptation $\gamma$, method $M$
\STATE {\bf Output:} Prediction interval $[L, U]$ with coverage $\geq 1-\alpha$
\STATE
\STATE {\bf Phase 1: Training}
\STATE Train HRF: $\hat{f}_{\text{HRF}} \leftarrow \text{HierarchicalRF}(\mathcal{D}_{\text{train}})$
\STATE Train Bayesian: $\theta \leftarrow \text{BayesianHRF}(\mathcal{D}_{\text{train}}, \hat{f}_{\text{HRF}})$
\STATE
\STATE {\bf Phase 2: Weighted Calibration Scores}
\FOR{each $(x_j, y_j, h_j, r_j) \in \mathcal{D}_{\text{calib}}$}
    \STATE $\hat{y}_j = \hat{f}_{\text{HRF}}(x_j, h_j, r_j)$; get $\sigma_j$ from posterior
    \STATE $S_j^{(w)} = |y_j - \hat{y}_j| / \max(\sigma_j^\gamma, \epsilon)$
\ENDFOR
\STATE
\STATE {\bf Phase 3: Hierarchical Conformal Calibration}
\IF{$M = \text{CDF pooling}$} 
    \STATE $\mathcal{S} = \{S_j^{(w)}\}_{j=1}^{n_{\text{calib}}}$
\ELSIF{$M = \text{Single subsampling}$} 
    \STATE $\mathcal{S} = \{S_{j_h}^{(w)} : h \in \text{hospitals}\}$
\ELSIF{$M = \text{Repeated subsampling}$} 
    \STATE $\hat{q} = \text{Median of } B \text{ subsampled quantiles}$
\ENDIF
\IF{$M \neq \text{Repeated subsampling}$}
    \STATE $\hat{q} = \text{Quantile}(\mathcal{S}, (1-\alpha)(|\mathcal{S}|+1)/|\mathcal{S}|)$
\ENDIF
\STATE
\STATE {\bf Phase 4: Test Prediction}
\STATE $\hat{y}_{\text{new}} = \hat{f}_{\text{HRF}}(x_{\text{new}}, h_{\text{new}}, r_{\text{new}})$
\STATE get $\sigma_{\text{new}}$ from Bayesian posterior
\STATE $w = \hat{q} \cdot \max(\sigma_{\text{new}}^\gamma, \epsilon)$
\STATE {\bf return} $[L, U] = [\hat{y}_{\text{new}} - w, \hat{y}_{\text{new}} + w]$
\end{algorithmic}
\label{ref:alg1}
\end{algorithm}

\subsection{Validation Metrics and Experimental Framework}
\label{sec:validation}

We used 5-fold stratified cross-validation on LOS quintiles with 64\%/16\%/20\% train/calibration/test splits. Missing values (2\%) were mean/median imputed, continuous variables were z-standardized. Hyperparameters were selected via grid search. HRF used three levels: patient (100 trees, depth 15), hospital (75, depth 12), region (50, depth 10). Bayesian models used MCMC (2 chains, 500 warmup, 250 post-warmup, target\_accept = 0.99) with convergence via $\hat{R}<1.01$ and ESS$>100$. 


Empirical coverage rate (target $\geq$95\% for $\alpha=0.05$) and conditional coverage within uncertainty quintiles served as primary metrics. Secondary metrics included interval width, uncertainty-error correlation (Pearson $r$), and adaptation index (width ratio for high versus low uncertainty). Robustness analyses examined 
performance across cross-validation folds, confidence levels (80\%, 90\%, 95\%, 99\%), hierarchical levels variation, and hospital subgroups.

\section{Experiments}

We analyzed 61,538 inpatient records from 3,793 hospitals across four U.S. census regions using the 2019 Healthcare Cost and Utilization Project (HCUP) National Inpatient Sample \cite{HCUP-NIS-Overview}, the largest publicly available all-payer inpatient database in the United States. All analyses used de-identified data consistent with HIPAA regulations \cite{HCUP-DUA}. 

The 18 predictor variables spanned three hierarchical levels: patient-level (demographics, clinical severity scores, diagnoses, procedures), hospital-level (bed size, teaching status, ownership), and regional-level (U.S. census region). This three-level hierarchical structure exhibited substantial clustering. Intraclass correlation coefficients revealed 12.5\% of LOS variance attributable to hospital-level factors, 40.8\% to regional factors, and 46.7\% to patient-level factors (total variance: 50.6 days²). This strong regional clustering (ICC = 0.408) demonstrates institutional variation that violates standard independence assumptions. LOS exhibited typical healthcare characteristics: median 3 days (IQR: 2-6 days), mean 4.9 days (SD: 7.1 days) with right skewness.

\begin{figure*}[t]
\centering
\begin{minipage}[t]{0.48\textwidth}
    \centering
    \includegraphics[width=\linewidth]{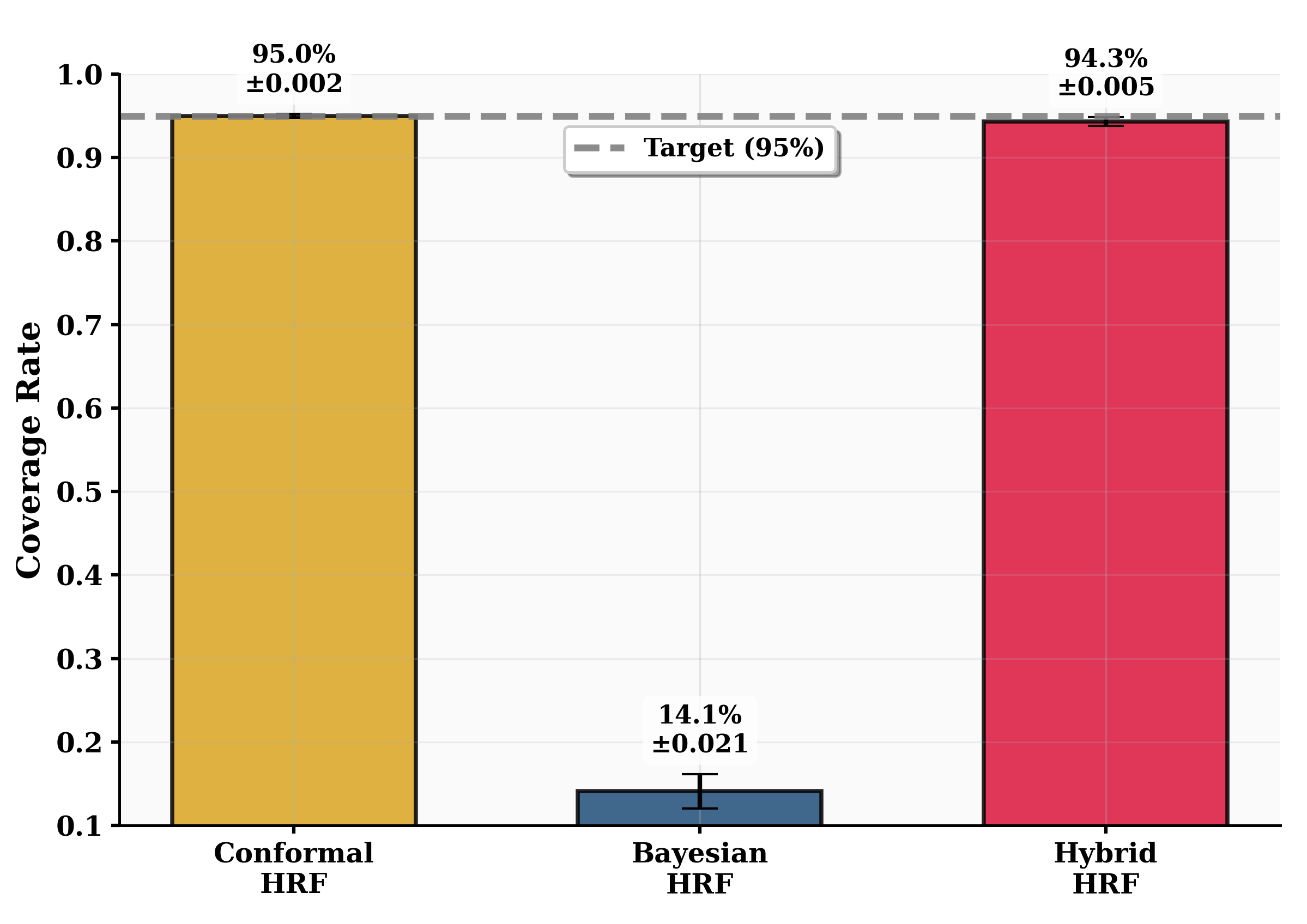}
    \centerline{(a) Overall coverage across five folds (mean $\pm$ SD).}
    \centerline{Dashed line: 95\% target.}
\end{minipage}
\hfill
\begin{minipage}[t]{0.48\textwidth}
    \centering
    \includegraphics[width=\linewidth]{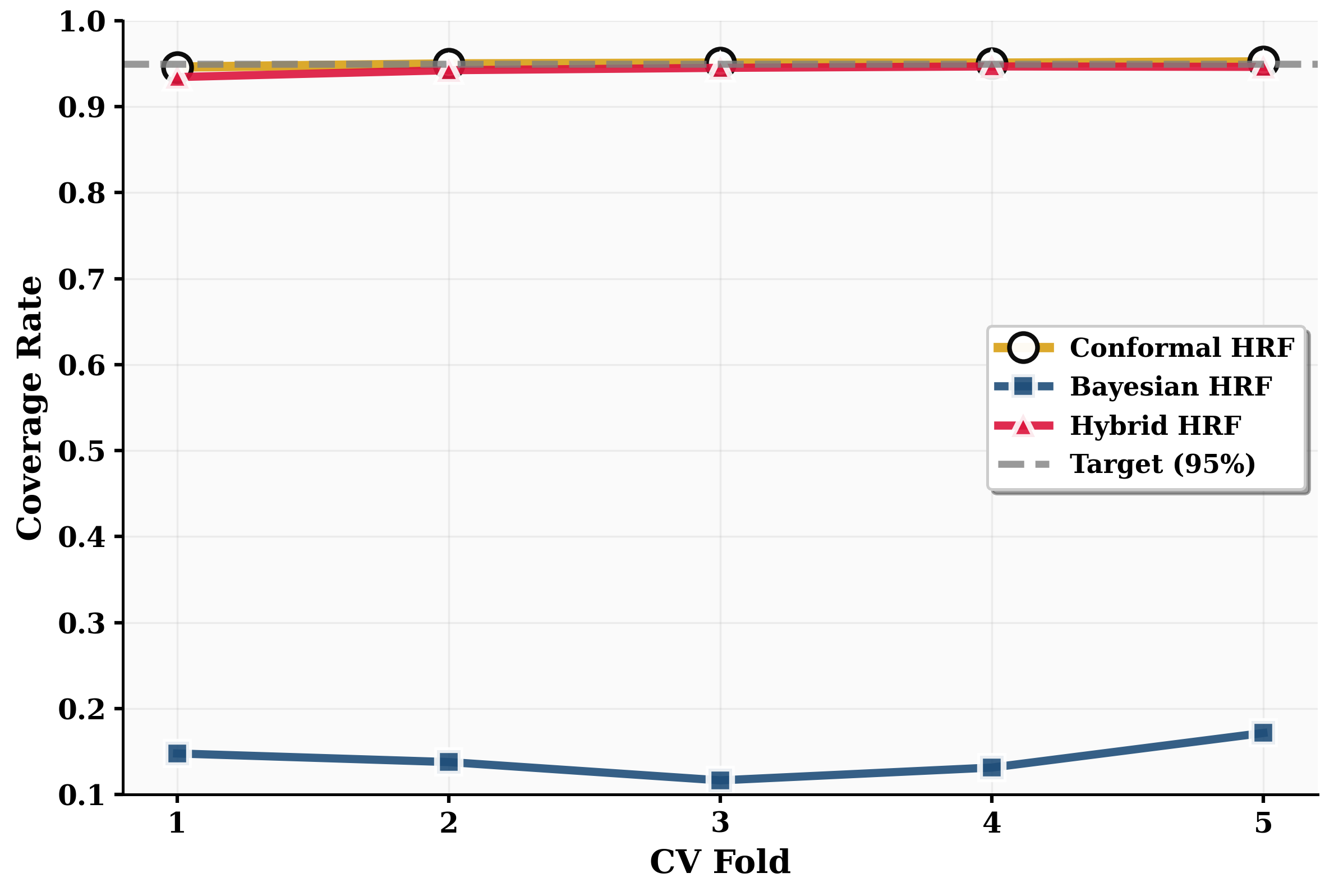}
    \centerline{(b) Coverage stability across folds. Conformal and Hybrid}
    \centerline{stable at $\approx$0.95; Bayesian varies 0.11--0.17.}
\end{minipage}
\caption{Coverage performance across methods}
\label{fig:coverage_performance}
\end{figure*}

\noindent \textbf{Coverage Validity.}\label{sec:coverage}
Across five-fold cross-validation, methods differed sharply in marginal coverage (Figure~\ref{fig:coverage_performance}A). Conformal HRF achieved 95.0\% ($\pm$0.2 pp), meeting the target (95\%). Hybrid HRF achieved 94.3\% ($\pm$0.5 pp), a modest 0.7 pp reduction. Bayesian HRF showed 14.1\% ($\pm$2.1 pp), indicating severe under-coverage. Coverage stability was high for Conformal HRF and Hybrid HRF (coefficients of variation (CV): 0.21\% and 0.53\%), with near-horizontal traces at $\sim$0.95 across folds (Figure~\ref{fig:coverage_performance}B). Bayesian HRF showed markedly greater fold-to-fold variability (CV: 14.9\%).

\begin{figure}[t]
\centering
\includegraphics[width=\columnwidth]{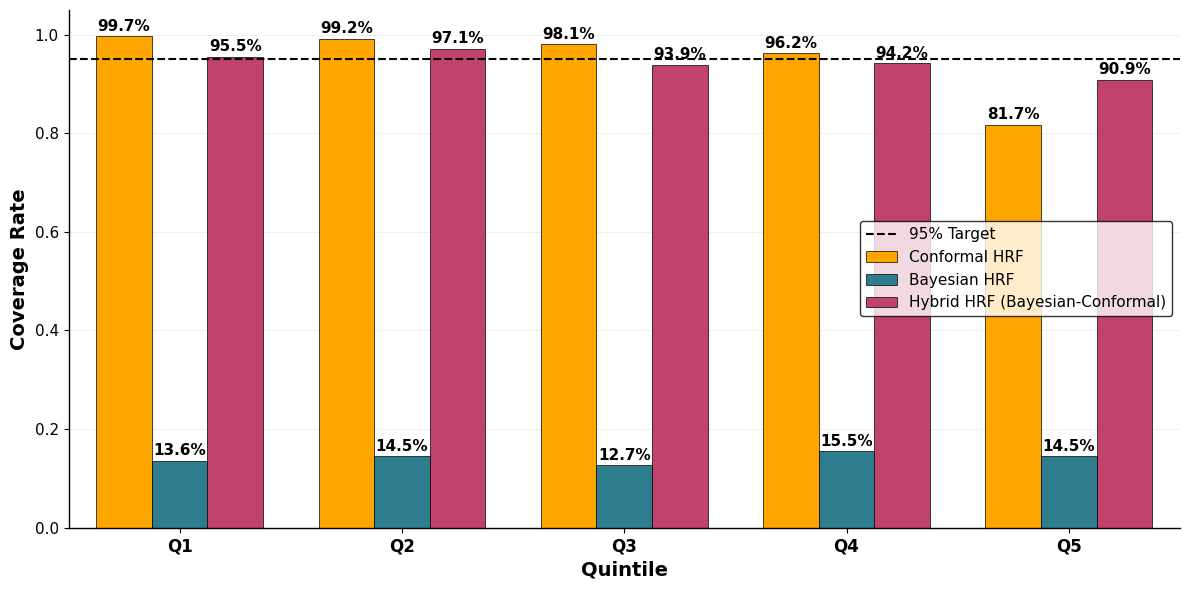}
\caption{Conditional coverage by uncertainty quintiles. (A) Bayesian HRF: uniformly poor (12.7\%--15.5\%). (B) Conformal HRF: Q1--Q4 near-perfect (96.2\%--99.7\%), Q5 under-coverage (81.7\%). (C) Hybrid HRF: Q1--Q4 strong (93.9\%--97.1\%), Q5 improved (90.9\%).}
\label{fig:conditional_coverage}
\end{figure}

Conditional coverage analysis stratified predictions by uncertainty quintiles (Q1 =lowest uncertainty/easiest cases, Q5 = highest uncertainty/hardest cases), revealing distinct patterns (Figure~\ref{fig:conditional_coverage}). Bayesian HRF showed uniformly poor coverage (12.7\%--15.5\%). Conformal HRF achieved near-perfect coverage in Q1--Q4 (96.2\%--99.7\%) but dropped sharply in Q5 (81.7\%, 
$-$13.3 pp from target). Hybrid HRF maintained strong coverage in Q1--Q4 (93.9\%--97.1\%) with smaller Q5 under-coverage (90.9\%, $-$4.1 pp from target), a 69\% reduction in coverage deficit for hardest cases versus Conformal HRF.

\noindent \textbf{Interval Efficiency and Adaptation.} Five-fold cross-validation revealed substantial differences in interval width characteristics (Table~\ref{tab:interval_width_analysis}). Bayesian HRF produced extremely narrow intervals (0.75 ± 0.02 days) with inadequate coverage reliability. Conformal HRF generated uniform intervals (16.32 ± 0.36 days, adaptation ratio = 1.00 by design). The Hybrid method achieved modestly narrower intervals (15.99 ± 0.59 days, 2.0\% reduction) while maintaining coverage validity, with meaningful adaptive sizing (adaptation ratio = 1.29, $P < 0.001$).

\begin{table*}[t]
\centering
\caption{Cross-Validated Interval Width Analysis}
\label{tab:interval_width_analysis}
\large
\begin{tabular}{lcccc}
\toprule
\textbf{Width Metric} & \textbf{Conformal} & \textbf{Bayesian} & \textbf{Hybrid} & \textbf{Difference$^\dagger$} \\
\midrule
Mean Width (days) & 16.32 $\pm$ 0.36 & 0.75 $\pm$ 0.02 & 15.99 $\pm$ 0.59 & -0.33 \\
Width Ratio vs Conformal & 1.00 $\pm$ 0.00 & 0.045 $\pm$ 0.001 & 0.98 $\pm$ 0.04 & -0.02 \\
Adaptation Ratio & 1.00 $\pm$ 0.00 & 1.01 $\pm$ 0.01 & 1.29 $\pm$ 0.01 & +0.29 \\
CV Stability (\%) & 2.2 & 2.8 & 3.7 & +1.5 \\
\bottomrule
\end{tabular}
\begin{minipage}{\textwidth}
\normalsize
\textit{Note:} $\pm$ = SD across 5 folds. $^\dagger$Difference = Hybrid vs Conformal. Adaptation Ratio = Q5/Q1 width ratio.
\end{minipage}
\end{table*}
Critically, the Hybrid method achieves efficiency through genuine optimization rather than coverage compromise. While the 2.0\% overall reduction appears modest, efficiency gains concentrate where they matter most: low-uncertainty cases (Q1) show 21\% narrower intervals (13.21 vs 16.71 days) enabling precise resource planning, while high-uncertainty cases (Q5) receive appropriately conservative widening (16.98 vs 15.99 days, +6\%) supporting safety-first protocols (Table~\ref{tab:quintile_width_analysis}). This risk-responsive sizing, quantified by the adaptation ratio of 1.29, enables uncertainty-stratified clinical workflows impossible with uniform conformal intervals.

\begin{table*}[t]
\centering
\caption{Interval Width by Uncertainty Quintile}
\label{tab:quintile_width_analysis}
\large
\begin{tabular}{lccccc}
\toprule
\textbf{Quintile} & \textbf{Mean $\sigma$} & \textbf{Conformal} & \textbf{Bayesian} & \textbf{Hybrid} & \textbf{Hyb/Conf} \\
\midrule
Q1 (Low)  & 0.168 & 16.71 $\pm$ 0.06 & 0.69 $\pm$ 0.01 & 13.21 $\pm$ 0.10 & 0.79 \\
Q2        & 0.176 & 16.24 $\pm$ 0.10 & 0.74 $\pm$ 0.01 & 14.86 $\pm$ 0.12 & 0.92 \\
Q3        & 0.179 & 16.06 $\pm$ 0.17 & 0.73 $\pm$ 0.02 & 15.14 $\pm$ 0.15 & 0.94 \\
Q4        & 0.186 & 16.62 $\pm$ 0.15 & 0.78 $\pm$ 0.01 & 16.48 $\pm$ 0.31 & 0.99 \\
Q5 (High) & 0.195 & 15.99 $\pm$ 0.10 & 0.81 $\pm$ 0.04 & 16.98 $\pm$ 1.38 & 1.06 \\
\bottomrule
\end{tabular}
\begin{minipage}{\textwidth}
\normalsize
\textit{Note:} Mean $\sigma$ = posterior predictive uncertainty (quintile formation). Width ratios vs Conformal baseline.
\end{minipage}
\end{table*}

\begin{figure*}[t]
\centering
\begin{minipage}[b]{0.32\textwidth}
    \centering
    \includegraphics[width=\textwidth]{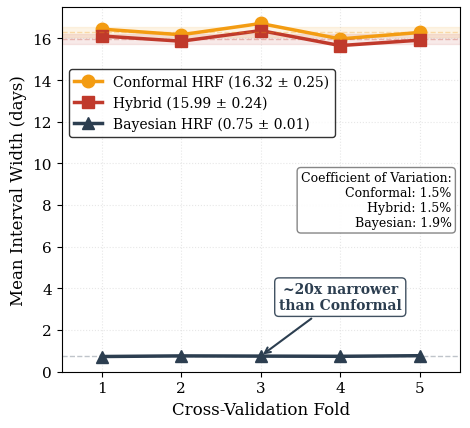}
    \vspace{2pt}  
    \centerline{(a) Width stability across}
    \centerline{methods and folds.}
\end{minipage}
\hfill
\begin{minipage}[b]{0.32\textwidth}
    \centering
    \includegraphics[width=\textwidth]{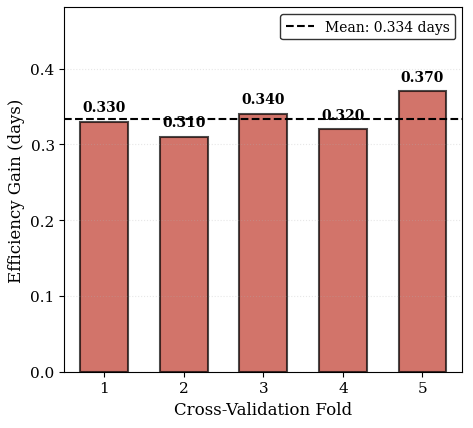}
    \vspace{2pt}
    \centerline{(b) Hybrid efficiency gains}
    \centerline{(all folds positive).}
\end{minipage}
\hfill
\begin{minipage}[b]{0.32\textwidth}
    \centering
    \includegraphics[width=\textwidth]{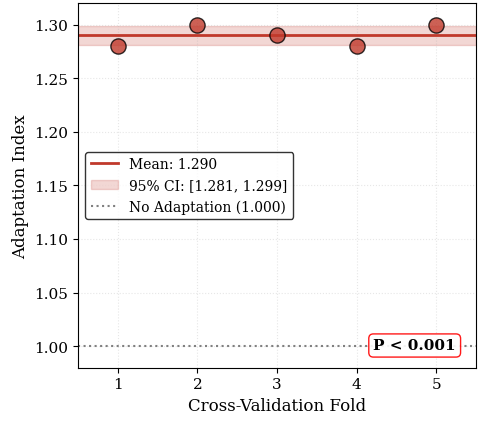}
    \vspace{2pt}
    \centerline{(c) Adaptation index}
    \centerline{consistency (P < 0.001).}
\end{minipage}
\caption{Cross-validation stability across 5 folds}
\label{fig:cv_stability}
\end{figure*}

Cross-validation stability confirmed robust performance (Figure~\ref{fig:cv_stability}). Conformal and Hybrid methods maintained consistent widths across folds (CV $< 2\%$), with the Hybrid method demonstrating positive efficiency gains in every fold (range: 0.31–0.37 days, no efficiency losses). The adaptation index remained stable (1.29 ± 0.01), with 95\% CI [1.28, 1.30] excluding the null hypothesis of uniform intervals (1.00, $P < 0.001$), confirming adaptive sizing as a genuine methodological characteristic rather than data artifact.

These results demonstrate that the Hybrid method provides genuine optimization, maintaining coverage validity while enabling risk-stratified resource allocation through adaptive interval sizing, a capability essential for uncertainty-guided clinical protocols that uniform conformal intervals cannot support.

\begin{figure*}[t]
\centering
\begin{minipage}[b]{0.48\textwidth}
    \centering
    \includegraphics[width=\textwidth]{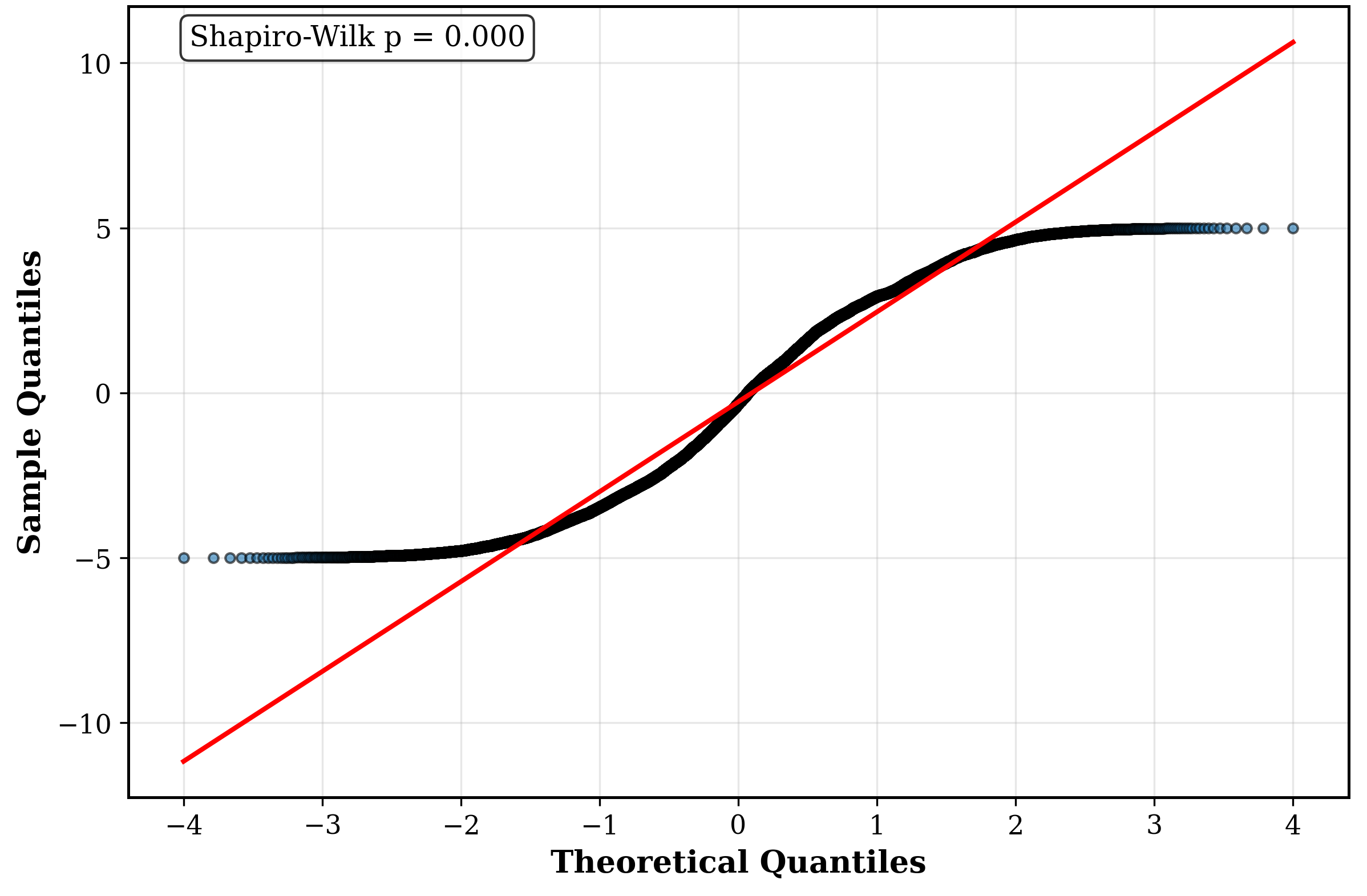}
    \centerline{(a) Raw Uncertainties}
\end{minipage}
\hfill
\begin{minipage}[b]{0.48\textwidth}
    \centering
    \includegraphics[width=\textwidth]{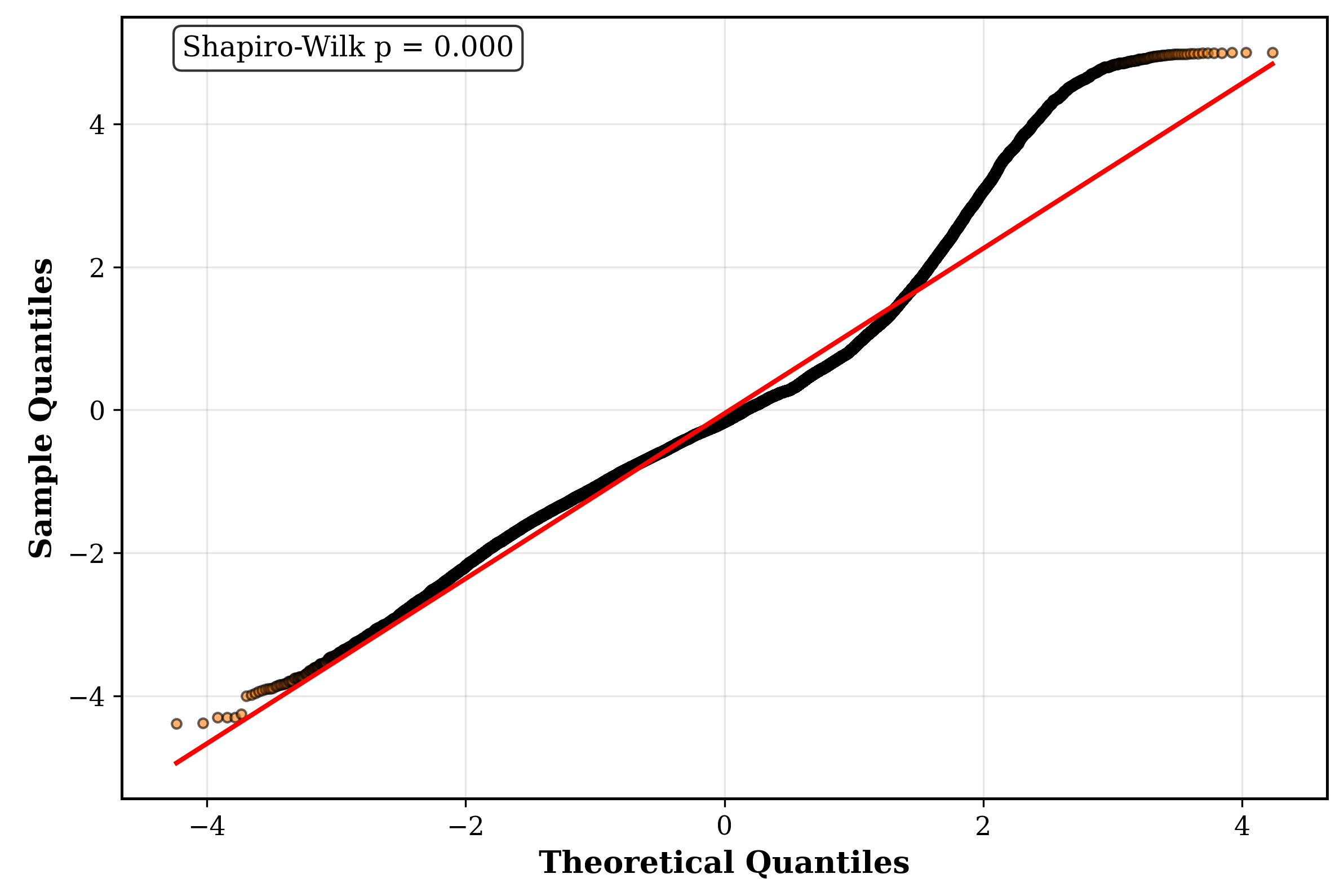}
    \centerline{(b) Calibrated Uncertainties}
\end{minipage}
\caption{QQ plots showing distributional improvement: (a) Raw uncertainties with systematic deviations, (b) Calibrated uncertainties approaching normality.}
\label{fig:qq_comparison}
\end{figure*}

\noindent \textbf{Uncertainty Calibration.} Raw Bayesian posterior uncertainties showed weak error correlation (Pearson $r = 0.203$, Spearman $\rho = 0.043$) and severe miscalibration. Isotonic regression calibration dramatically improved uncertainty quality (Table~\ref{tab:uncertainty_calibration}), increasing Pearson $r$ to 0.476 (+134\%), Spearman $\rho$ to 0.243 (+465\%), improving calibration slope from 91.4 to 0.926, reducing Expected Calibration Error from 
2.52 to 0.037 days (98.5\% reduction), and mean uncertainty from 0.19 to 2.71 days (+1326\%). QQ plots (Figure~\ref{fig:qq_comparison}) confirmed distributional improvement approaching normality.

\begin{table}[t]
\centering
\caption{Isotonic Calibration Impact on Uncertainty Quality}
\label{tab:uncertainty_calibration}
\begin{tabular}{@{}l@{\hspace{3pt}}c@{\hspace{3pt}}c@{\hspace{3pt}}c@{\hspace{3pt}}c@{\hspace{3pt}}c@{}}
\toprule
\textbf{State} & \textbf{Pearson} & \textbf{Spearman} & \textbf{Slope} & \textbf{ECE} & \textbf{Mean} \\
 & $r$ & $\rho$ & $\beta$ & & $\sigma$ \\
\midrule
Raw & 0.203 & 0.043 & 91.4 & 2.52 & 0.19 \\
Calibrated & 0.476 & 0.243 & 0.926 & 0.037 & 2.71 \\
\midrule
Improv. & $+134\%$ & $+465\%$ & $-98.9\%$ & $-98.5\%$ & $+1326\%$ \\
\bottomrule
\end{tabular}
\fontsize{7.5pt}{9pt}\selectfont
\end{table}

Despite well-calibrated uncertainties, coverage performance differed dramatically (Table~\ref{tab:coverage_performance}). Bayesian HRF achieved only 14.1\% coverage with narrow intervals (0.75 days), while Hybrid HRF achieved 94.3\% coverage with appropriate widths (15.99 days). Identical CRPS (2.04) confirms equivalent discrimination. Superior Winkler scores (35.5 vs 95.0) demonstrate reliable coverage outweighs narrow intervals, and this validates the framework's design.

\begin{table}[t]
\centering
\caption{Coverage Performance with Calibrated Uncertainties}
\label{tab:coverage_performance}
\normalsize
\begin{tabular}{@{}lcccc@{}}
\toprule
\textbf{Method} & \textbf{Coverage} & \textbf{Width} & \textbf{CRPS} & \textbf{Winkler} \\
 & (\%) & (days) & & \textbf{Score} \\
\midrule
BayHRF & 14.1 & 0.75 & 2.04 & 95.02 \\
HybridHRF & 94.3 & 15.99 & 2.04 & 35.52 \\
\bottomrule
\end{tabular}
\end{table}

\noindent \textbf{Clinical Interpretation.} The 94.3\% coverage rate means hospital planners can expect approximately 943 of 1,000 admissions to have actual LOS within predicted bounds, providing quantifiable reliability for capacity management. Knowing that approximately 57 cases per 1,000 will exceed intervals allows advance planning for escalation capacity, allocating overflow beds, scheduling discharge planning resources, or coordinating with post, acute care networks, converting unpredictable demand into managed, budgeted contingency capacity.
%
%
High-confidence predictions (Q1-Q2) receive narrower intervals than conformal methods (Q1: 13.21 vs 16.71 days), enabling 1.4-3.5 day buffer reductions with standard monitoring. High-uncertainty cases (Q5) receive appropriately conservative intervals, triggering enhanced oversight. 




\noindent \textbf{Robustness and Generalizability.} Comprehensive robustness analyses confirmed stable performance across multiple dimensions (Table~\ref{tab:robustness_comprehensive}). Predictive accuracy was equivalent across methods (RMSE: 6.0--6.4 days, Panel A). Conformal and Hybrid methods appropriately tracked nominal targets across confidence levels (80--99\%), while Bayesian under-covered (Panel B). Coverage remained stable across all hospital characteristics (bed size, ownership, location, and geographic region) with Conformal/Hybrid achieving 93.5--95.6\%  versus Bayesian's 11.0--40.6\% (Panel C). Alternative hierarchy specifications showed minimal 
impact on Hybrid performance (coverage: 94.2--94.8\%, width: 15.80--16.30 days, Panel D).

\begin{table*}[h!]
\centering
\caption{Comprehensive Robustness Analysis Across Multiple Dimensions}
\label{tab:robustness_comprehensive}
\small
\begin{tabular}{@{}llccc@{}}
\toprule
\textbf{Dimension} & \textbf{Condition} & \textbf{Conformal} & \textbf{Bayesian} & \textbf{Hybrid} \\
\midrule
\multicolumn{5}{c}{\textbf{A. Predictive Accuracy (RMSE, days)}} \\
\midrule
& Cross-validation (5-fold, mean ± SD) & 6.05 ± 0.59 & 6.39 ± 0.60 & 6.01 ± 0.65 \\
\midrule
\multicolumn{5}{c}{\textbf{B. Confidence Level Sensitivity (Coverage \%, mean ± SD)}} \\
\midrule
& $\alpha=0.01$ (99\% target) & 99.1 ± 0.1 & 22.1 ± 2.1 & 99.0 ± 0.2 \\
& $\alpha=0.05$ (95\% target) & 95.0 ± 0.4 & 14.1 ± 2.1 & 94.4 ± 0.3 \\
& $\alpha=0.10$ (90\% target) & 90.0 ± 0.6 & 12.0 ± 1.0 & 88.9 ± 0.7 \\
& $\alpha=0.20$ (80\% target) & 80.0 ± 0.6 & 8.0 ± 0.6 & 79.0 ± 1.6 \\
\midrule
\multicolumn{5}{c}{\textbf{C. Hospital Characteristics (Coverage \%, mean ± SD)}} \\
\midrule
\multicolumn{1}{l}{\textit{Bed Size}} & & & & \\
& Small (n=13,534) & 95.1 ± 1.0 & 11.0 ± 1.9 & 95.3 ± 1.0 \\
& Medium (n=17,735) & 95.3 ± 0.2 & 13.9 ± 2.4 & 95.2 ± 0.1 \\
& Large (n=30,269) & 94.8 ± 0.6 & 15.4 ± 1.7 & 94.3 ± 1.1 \\
\addlinespace
\multicolumn{1}{l}{\textit{Ownership}} & & & & \\
& Public (n=7,023) & 93.5 ± 2.2 & 12.7 ± 1.4 & 93.5 ± 3.0 \\
& Private Non-profit (n=45,896) & 95.0 ± 0.3 & 15.8 ± 2.2 & 94.7 ± 0.6 \\
& Private For-profit (n=8,619) & 94.9 ± 1.7 & 13.6 ± 1.8 & 94.0 ± 0.8 \\
\addlinespace
\multicolumn{1}{l}{\textit{Location}} & & & & \\
& Rural (n=5,259) & 94.8 ± 2.1 & 14.2 ± 0.8 & 94.7 ± 1.3 \\
& Urban Non-teaching (n=10,868) & 95.1 ± 0.3 & 15.6 ± 4.1 & 94.8 ± 1.3 \\
& Urban Teaching (n=45,411) & 94.7 ± 0.7 & 13.7 ± 1.4 & 94.8 ± 1.0 \\
\addlinespace
\multicolumn{1}{l}{\textit{Region}} & & & & \\
& Northeast (n=11,223) & 95.2 ± 1.1 & 37.9 ± 4.4 & 95.4 ± 0.8 \\
& Midwest (n=13,654) & 95.6 ± 0.1 & 38.7 ± 1.3 & 94.6 ± 0.6 \\
& South (n=24,643) & 94.9 ± 0.2 & 38.7 ± 0.5 & 94.1 ± 0.5 \\
& West (n=12,018) & 95.0 ± 0.4 & 40.6 ± 3.3 & 94.7 ± 1.1 \\
\midrule
\multicolumn{5}{c}{\textbf{D. Hierarchy Specification Sensitivity (Hybrid HRF)}} \\
\midrule
& 3-Level (Patient→Hospital→Region) & \multicolumn{3}{c}{94.3 ± 0.1\% / 15.99 ± 0.59 days} \\
& 2-Level (Patient→Hospital) & \multicolumn{3}{c}{94.8 ± 0.3\% / 16.30 ± 0.34 days} \\
& 2-Level (Patient→Region) & \multicolumn{3}{c}{94.2 ± 0.4\% / 15.80 ± 0.42 days} \\
\bottomrule
\end{tabular}
\vspace{0.1cm}
\par\noindent\footnotesize
Coverage and width values shown as mean ± SD. Panel A: Point prediction equivalence across methods. Panel B: Target tracking for different confidence levels. Panel C: Performance across all hospital subgroups. Panel D: Coverage/width for different hierarchical structures.
\end{table*}

\section{Discussion}


This study demonstrates that hybrid hierarchical (Bayesian Conformal HRF) uncertainty quantification achieves reliable coverage while providing adaptive precision for healthcare prediction. Using data from 61,538 patients from 3,793 hospitals in four U.S. regions, our framework attained 94.3\% coverage on held-out test sets, delivering 21\% narrower intervals for low-uncertainty cases and conservative widening for high-uncertainty cases (adaptation ratio: 1.29, 95\% CI [1.28, 1.30]). Performance remained stable in hospital characteristics, validation folds, and confidence levels.

The primary contribution is demonstrating that distribution-free calibration is essential for safety-critical applications regardless of underlying uncertainty model quality. Bayesian intervals alone achieved only 14.1\% coverage despite identical well-calibrated uncertainty discrimination as the hybrid method (CRPS: 2.04 for both), showing that uncertainty calibration cannot substitute for formal coverage validity. This validates our modular design: Bayesian hierarchical models for adaptive weighting, conformal calibration for coverage validity. Our approach differs from existing adaptive conformal methods. While locally-weighted conformal prediction and quantile regression forests also provide adaptive intervals, our framework offers computational efficiency through offline Bayesian training rather than per-point reweighting, modular separation of coverage guarantees from uncertainty estimation, and hierarchical structure naturally suited to clustered healthcare data. However, systematic comparison to these methods remains important future work.

\noindent \textbf{Limitations and Mitigation.} The primary limitation is under-coverage in the highest-uncertainty quintile (Q5: 90.9\% vs 95\% target, -4.1 percentage points), affecting patients where reliable bounds matter most. This occurs because a single global conformal multiplier cannot fully accommodate heteroscedastic error patterns despite improved uncertainty discrimination (Pearson r: 0.203 → 0.476 after calibration). While conformal theory guarantees marginal coverage under exchangeability, conditional coverage within uncertainty-defined subsets is not ensured.

Mitigation strategies include uncertainty-stratified calibration using separate conformal quantiles per quintile, conservative widening for high-uncertainty cases (e.g., 1.2× multiplier above 75th percentile), and enhanced clinical oversight requiring senior review for flagged cases. Deployment decisions must weigh coverage shortfall against clinical risk tolerance. Settings prioritizing absolute safety may require the conservatism of standard conformal methods until Q5 coverage is resolved, while general medical-surgical populations can mitigate modest under-coverage through enhanced oversight protocols.

\section{Conclusion}

This study demonstrates that combining Bayesian hierarchical uncertainty quantification with distribution-free conformal calibration achieves reliable, adaptive prediction intervals for clustered healthcare data. Evaluating on 61,538 patients from 3,793 hospitals, the framework attained near-target coverage (94.3\% vs 95\% target) on held-out test data with meaningful adaptive precision, while revealing that well-calibrated uncertainty models cannot substitute for formal coverage validity in safety-critical applications. Prospective validation in diverse clinical settings represents the essential next step toward deployment.

\section*{Acknowledgment}
This material is based upon work supported by the National Science Foundation under Award No. DGE-2125362. Any opinions, findings, and conclusions or recommendations expressed in this material are those of the author(s) and do not necessarily reflect the views of the National Science Foundation.

\bibliography{main}

@article{anandalingam1992hierarchical,
  title={Hierarchical optimization: An introduction},
  author={Anandalingam, Gnana and Friesz, Terry L},
  journal={Annals of operations research},
  volume={34},
  number={1},
  pages={1--11},
  year={1992},
  publisher={Springer}
}

@article{wong1996hierarchical,
  title={Hierarchical control of discrete-event systems},
  author={Wong, Kai C and Wonham, Walter Murray},
  journal={Discrete Event Dynamic Systems},
  volume={6},
  number={3},
  pages={241--273},
  year={1996},
  publisher={Springer}
}

@article{baheri2025multilevel,
  title={Multilevel constrained bandits: A hierarchical upper confidence bound approach with safety guarantees},
  author={Baheri, Ali},
  journal={Mathematics},
  volume={13},
  number={1},
  pages={149},
  year={2025},
  publisher={MDPI}
}

@article{bafandeh2018comparative,
  title={A comparative assessment of hierarchical control structures for spatiotemporally-varying systems, with application to airborne wind energy},
  author={Bafandeh, Alireza and Bin-Karim, Shamir and Baheri, Ali and Vermillion, Christopher},
  journal={Control Engineering Practice},
  volume={74},
  pages={71--83},
  year={2018},
  publisher={Elsevier}
}

@article{shahbazi2025hierarchical,
  title={Hierarchical data modeling: A systematic comparison of statistical, tree-based, and neural network approaches},
  author={Shahbazi, Marzieh Amiri and Azadeh-Fard, Nasibeh},
  journal={Machine Learning with Applications},
  pages={100688},
  year={2025},
  publisher={Elsevier}
}

@article{cressie2009accounting,
  title={Accounting for uncertainty in ecological analysis: the strengths and limitations of hierarchical statistical modeling},
  author={Cressie, Noel and Calder, Catherine A and Clark, James S and Hoef, Jay M Ver and Wikle, Christopher K},
  journal={Ecological Applications},
  volume={19},
  number={3},
  pages={553--570},
  year={2009},
  publisher={Wiley Online Library}
}

@article{mentch2016quantifying,
  title={Quantifying uncertainty in random forests via confidence intervals and hypothesis tests},
  author={Mentch, Lucas and Hooker, Giles},
  journal={Journal of Machine Learning Research},
  volume={17},
  number={26},
  pages={1--41},
  year={2016}
}

@article{wager2018estimation,
  title={Estimation and inference of heterogeneous treatment effects using random forests},
  author={Wager, Stefan and Athey, Susan},
  journal={Journal of the American Statistical Association},
  volume={113},
  number={523},
  pages={1228--1242},
  year={2018},
  publisher={Taylor \& Francis}
}

@article{vovk2009line,
  title={Linearly time efficient nonconformity measure for conformal prediction},
  author={Vovk, Vladimir and Nouretdinov, Ilia and Gammerman, Alex},
  journal={Annals of Mathematics and Artificial Intelligence},
  volume={56},
  number={1},
  pages={83--99},
  year={2009},
  publisher={Springer}
}

@article{lei2018distribution,
  title={Distribution-free predictive inference for regression},
  author={Lei, Jing and G’Sell, Max and Rinaldo, Alessandro and Tibshirani, Ryan J and Wasserman, Larry},
  journal={Journal of the American Statistical Association},
  volume={113},
  number={523},
  pages={1094--1111},
  year={2018},
  publisher={Taylor & Francis}
}

@article{chipman2010bart,
  title={BART: Bayesian additive regression trees},
  author={Chipman, Hugh A and George, Edward I and McCulloch, Robert E},
  journal={The Annals of Applied Statistics},
  volume={4},
  number={1},
  pages={266--298},
  year={2010},
  publisher={Institute of Mathematical Statistics}
}

@article{linero2018bayesian,
  title={Bayesian regression trees for high-dimensional prediction and variable selection},
  author={Linero, Antonio R},
  journal={Journal of the American Statistical Association},
  volume={113},
  number={522},
  pages={626--636},
  year={2018},
  publisher={Taylor & Francis}
}

@article{gelman2013bayesian,
  title={Bayesian data analysis},
  author={Gelman, Andrew and Carlin, John B and Stern, Hal S and Dunson, David B and Vehtari, Aki and Rubin, Donald B},
  journal={CRC press},
  year={2013}
}

@article{meehinkong2025coverforest,
  title={coverforest: Conformal Predictions with Random Forest in Python},
  author={Meehinkong, Panisara and Ponnoprat, Donlapark},
  journal={arXiv preprint arXiv:2501.14570},
  year={2025}
}

@article{tyralis2024review,
  title={A review of predictive uncertainty estimation with machine learning},
  author={Tyralis, Hristos and Papacharalampous, Georgia},
  journal={Artificial Intelligence Review},
  volume={57},
  number={4},
  pages={94},
  year={2024},
  publisher={Springer}
}

@article{pellagatti2021generalized,
  title={Generalized mixed-effects random forest: A flexible approach to predict university student dropout},
  author={Pellagatti, Massimo and Masci, Chiara and Ieva, Francesca and Paganoni, Anna M},
  journal={Statistical Analysis and Data Mining: The ASA Data Science Journal},
  volume={14},
  number={3},
  pages={241--257},
  year={2021},
  publisher={Wiley Online Library}
}

@article{hajjem2010mixed,
  title={Mixed effects trees and forests for clustered data},
  author={Hajjem, Ahlem},
  journal={ProQuest LLC: University of Montreal},
  year={2010}
}

@article{sela2012re,
  title={RE-EM trees: a data mining approach for longitudinal and clustered data},
  author={Sela, Rebecca J and Simonoff, Jeffrey S},
  journal={Machine learning},
  volume={86},
  pages={169--207},
  year={2012},
  publisher={Springer}
}

@article{angelopoulos2023conformal,
  title={Conformal prediction: A gentle introduction},
  author={Angelopoulos, Anastasios N and Bates, Stephen and others},
  journal={Foundations and Trends{\textregistered} in Machine Learning},
  volume={16},
  number={4},
  pages={494--591},
  year={2023},
  publisher={Now Publishers, Inc.}
}

@inproceedings{baheri2023safety,
  title={Safety validation of learning-based autonomous systems: A multi-fidelity approach},
  author={Baheri, Ali},
  booktitle={Proceedings of the AAAI Conference on Artificial Intelligence},
  volume={37},
  number={13},
  pages={15432--15432},
  year={2023}
}

@article{dunn2023distribution,
  title={Distribution-free prediction sets for two-layer hierarchical models},
  author={Dunn, Robin and Wasserman, Larry and Ramdas, Aaditya},
  journal={Journal of the American Statistical Association},
  volume={118},
  number={544},
  pages={2491--2502},
  year={2023},
  publisher={Taylor \& Francis}
}

@article{barber2023conformal,
  title={Conformal prediction beyond exchangeability},
  author={Barber, Rina Foygel and Candes, Emmanuel J and Ramdas, Aaditya and Tibshirani, Ryan J},
  journal={The Annals of Statistics},
  volume={51},
  number={2},
  pages={816--845},
  year={2023},
  publisher={Institute of Mathematical Statistics}
}

@article{tan2019bayesian,
  title={Bayesian additive regression trees and the General BART model},
  author={Tan, Yaoyuan Vincent and Roy, Jason},
  journal={Statistics in medicine},
  volume={38},
  number={25},
  pages={5048--5069},
  year={2019},
  publisher={Wiley Online Library}
}

@article{gelman2006prior,
  title={Prior distributions for variance parameters in hierarchical models (comment on article by Browne and Draper)},
  author={Gelman, Andrew},
  year={2006}
}

@article{carlin1999assessing,
  title={Assessing environmental justice using Bayesian hierarchical models: two case studies.},
  author={Carlin, Bradley P and Xia, Hong},
  journal={Journal of Exposure Analysis \& Environmental Epidemiology},
  volume={9},
  number={1},
  year={1999}
}

@article{fong2021conformal,
  title={Conformal bayesian computation},
  author={Fong, Edwin and Holmes, Chris C},
  journal={Advances in Neural Information Processing Systems},
  volume={34},
  pages={18268--18279},
  year={2021}
}

@article{ekmekci2025conformalized,
  title={Conformalized Generative Bayesian Imaging: An Uncertainty Quantification Framework for Computational Imaging},
  author={Ekmekci, Canberk and Cetin, Mujdat},
  journal={arXiv preprint arXiv:2504.07696},
  year={2025}
}

@article{podina2024conformalized,
  title={Conformalized physics-informed neural networks},
  author={Podina, Lena and Rad, Mahdi Torabi and Kohandel, Mohammad},
  journal={arXiv preprint arXiv:2405.08111},
  year={2024}
}

@inproceedings{stanton2023bayesian,
  title={Bayesian optimization with conformal prediction sets},
  author={Stanton, Samuel and Maddox, Wesley and Wilson, Andrew Gordon},
  booktitle={International Conference on Artificial Intelligence and Statistics},
  pages={959--986},
  year={2023},
  organization={PMLR}
}

@article{zhang2024bayesian,
  title={Bayesian optimization with formal safety guarantees via online conformal prediction},
  author={Zhang, Yunchuan and Park, Sangwoo and Simeone, Osvaldo},
  journal={IEEE Journal of Selected Topics in Signal Processing},
  year={2024},
  publisher={IEEE}
}

@article{ovadia2019can,
  title={Can you trust your model's uncertainty? evaluating predictive uncertainty under dataset shift},
  author={Ovadia, Yaniv and Fertig, Emily and Ren, Jie and Nado, Zachary and Sculley, David and Nowozin, Sebastian and Dillon, Joshua and Lakshminarayanan, Balaji and Snoek, Jasper},
  journal={Advances in neural information processing systems},
  volume={32},
  year={2019}
}

@article{amodei2016concrete,
  title={Concrete problems in AI safety},
  author={Amodei, Dario and Olah, Chris and Steinhardt, Jacob and Christiano, Paul and Schulman, John and Man{\'e}, Dan},
  journal={arXiv preprint arXiv:1606.06565},
  year={2016}
}

@inproceedings{guo2017calibration,
  title={On calibration of modern neural networks},
  author={Guo, Chuan and Pleiss, Geoff and Sun, Yu and Weinberger, Kilian Q},
  booktitle={International conference on machine learning},
  pages={1321--1330},
  year={2017},
  organization={PMLR}
}

@book{goldstein2011multilevel,
  title={Multilevel statistical models},
  author={Goldstein, Harvey},
  year={2011},
  publisher={John Wiley \& Sons}
}

@article{snijders2011multilevel,
  title={Multilevel analysis: An introduction to basic and advanced multilevel modeling},
  author={Snijders, Tom AB and Bosker, Roel},
  year={2011},
  publisher={sage}
}

@article{rajkomar2018scalable,
  title={Scalable and accurate deep learning with electronic health records},
  author={Rajkomar, Alvin and Oren, Eyal and Chen, Kai and Dai, Andrew M and Hajaj, Nissan and Hardt, Michaela and Liu, Peter J and Liu, Xiaobing and Marcus, Jake and Sun, Mimi and others},
  journal={NPJ digital medicine},
  volume={1},
  number={1},
  pages={18},
  year={2018},
  publisher={Nature Publishing Group UK London}
}

@article{stone2022systematic,
  title={A systematic review of the prediction of hospital length of stay: Towards a unified framework},
  author={Stone, Kieran and Zwiggelaar, Reyer and Jones, Phil and Mac Parthal{\'a}in, Neil},
  journal={PLOS digital health},
  volume={1},
  number={4},
  pages={e0000017},
  year={2022},
  publisher={Public Library of Science}
}

@book{murphy2012machine,
  title={Machine learning: a probabilistic perspective},
  author={Murphy, Kevin P},
  year={2012},
  publisher={MIT press}
}

@article{pratola2020heteroscedastic,
  title={Heteroscedastic BART via multiplicative regression trees},
  author={Pratola, Matthew T and Chipman, Hugh A and George, Edward I and McCulloch, Robert E},
  journal={Journal of Computational and Graphical Statistics},
  volume={29},
  number={2},
  pages={405--417},
  year={2020},
  publisher={Taylor \& Francis}
}

@book{raudenbush2002hierarchical,
  title={Hierarchical linear models: Applications and data analysis methods},
  author={Raudenbush, Stephen W and Bryk, Anthony S},
  volume={1},
  year={2002},
  publisher={sage}
}

@article{vehtari2017practical,
  title={Practical Bayesian model evaluation using leave-one-out cross-validation and WAIC},
  author={Vehtari, Aki and Gelman, Andrew and Gabry, Jonah},
  journal={Statistics and computing},
  volume={27},
  pages={1413--1432},
  year={2017},
  publisher={Springer}
}

@article{beam2018big,
  title={Big data and machine learning in health care},
  author={Beam, Andrew L and Kohane, Isaac S},
  journal={Jama},
  volume={319},
  number={13},
  pages={1317--1318},
  year={2018},
  publisher={American Medical Association}
}

@article{shafer2008tutorial,
  title={A tutorial on conformal prediction.},
  author={Shafer, Glenn and Vovk, Vladimir},
  journal={Journal of Machine Learning Research},
  volume={9},
  number={3},
  year={2008}
}

@article{tibshirani2019conformal,
  title={Conformal prediction under covariate shift},
  author={Tibshirani, Ryan J and Foygel Barber, Rina and Candes, Emmanuel and Ramdas, Aaditya},
  journal={Advances in neural information processing systems},
  volume={32},
  year={2019}
}

@article{breiman2001random,
  title={Random forests},
  author={Breiman, Leo},
  journal={Machine learning},
  volume={45},
  pages={5--32},
  year={2001},
  publisher={Springer}
}

@article{betancourt2017conceptual,
  title={A conceptual introduction to Hamiltonian Monte Carlo},
  author={Betancourt, Michael},
  journal={arXiv preprint arXiv:1701.02434},
  year={2017}
}

@inproceedings{varshney2016engineering,
  title={Engineering safety in machine learning},
  author={Varshney, Kush R},
  booktitle={2016 Information Theory and Applications Workshop (ITA)},
  pages={1--5},
  year={2016},
  organization={IEEE}
}

@article{principato2024conformal,
  title={Conformal prediction for hierarchical data},
  author={Principato, Guillaume and Stoltz, Gilles and Amara-Ouali, Yvenn and Goude, Yannig and Hamrouche, Bachir and Poggi, Jean-Michel},
  journal={arXiv preprint arXiv:2411.13479},
  year={2024}
}

@misc{HCUP-DUA,
  author = {{Agency for Healthcare Research and Quality (AHRQ)}},
  title = {{Nationwide Data Use Agreement - HCUP}},
  year = {2024},
  note = {Accessed July 22, 2025},
  howpublished = {\url{https://hcup-us.ahrq.gov/team/NationwideDUA.jsp}}
}

@misc{HCUP-NIS-Overview,
  author = {{Agency for Healthcare Research and Quality (AHRQ)}},
  title = {{HCUP-US NIS Overview}},
  year = {2025},
  note = {Accessed July 22, 2025},
  howpublished = {\url{https://hcup-us.ahrq.gov/nisoverview.jsp}}
}

@article{baheri2025conformal,
  title={Conformal prediction across scales: Finite-sample coverage with hierarchical efficiency},
  author={Baheri, Ali and Shahbazi, Marzieh Amiri},
  journal={Results in Applied Mathematics},
  volume={26},
  pages={100589},
  year={2025},
  publisher={Elsevier}
}

@article{mccarthy2021improving,
  title={Improving hospital length of stay prediction through machine learning},
  author={McCarthy, Katherine and others},
  journal={Journal of Healthcare Management},
  volume={66},
  number={4},
  pages={272--285},
  year={2021}
}

@article{harris2018bed,
  title={Bed management and length of stay},
  author={Harris, Sarah L and others},
  journal={International Journal of Health Care Quality Assurance},
  volume={31},
  number={8},
  pages={1014--1027},
  year={2018}
}

@article{iezzoni2012risk,
  title={Risk adjustment for measuring health care outcomes},
  author={Iezzoni, Lisa I},
  journal={Health Administration Press},
  year={2012}
}

@article{dimick2013hospital,
  title={Hospital volume and surgical outcomes: is more always better?},
  author={Dimick, Justin B and others},
  journal={New England Journal of Medicine},
  volume={369},
  pages={1073--1075},
  year={2013}
}

@article{freeland2013length,
  title={Length of stay determinants in hospital medicine},
  author={Freeland, Kenneth N and others},
  journal={American Journal of Medical Quality},
  volume={28},
  number={5},
  pages={357--365},
  year={2013}
}

@article{kaboli2012associations,
  title={Associations between reduced hospital length of stay and 30-day readmission rate and mortality},
  author={Kaboli, Peter J and others},
  journal={Annals of Internal Medicine},
  volume={157},
  number={12},
  pages={837--845},
  year={2012}
}

@article{arthur2020planning,
  title={Planning hospital capacity to save lives during the COVID-19 pandemic and beyond},
  author={Arthur, Richard and others},
  journal={European Journal of Operational Research},
  volume={295},
  number={3},
  pages={1202--1214},
  year={2022}
}

@article{helm2011improving,
  title={Improving hospital efficiency and patient flow},
  author={Helm, Joanna E and others},
  journal={Production and Operations Management},
  volume={20},
  number={3},
  pages={385--397},
  year={2011}
}

@article{proudlove2003can,
  title={Can good bed management solve the overcrowding in accident and emergency departments?},
  author={Proudlove, Nathan C and others},
  journal={Emergency Medicine Journal},
  volume={20},
  number={2},
  pages={149--155},
  year={2003}
}

@article{liu2020reporting,
  title={Reporting guidelines for clinical trial reports for interventions involving artificial intelligence},
  author={Liu, Xiaoxuan and others},
  journal={The Lancet Digital Health},
  volume={2},
  number={10},
  pages={e537--e548},
  year={2020}
}

@article{werner2023explainable,
  title={Explainable hierarchical clustering for patient subtyping and risk prediction},
  author={Werner, Enrico and Clark, Jeffrey N and Hepburn, Alexander and Bhamber, Ranjeet S and Ambler, Michael and Bourdeaux, Christopher P and McWilliams, Christopher J and Santos-Rodriguez, Raul},
  journal={Experimental Biology and Medicine},
  volume={248},
  number={24},
  pages={2547--2559},
  year={2023},
  publisher={SAGE Publications}
}

@article{bollmann2023accounting,
  title={Accounting for clustering in automated variable selection using hospital data: a comparison of different LASSO approaches},
  author={Bollmann, Stella and Groll, Andreas and Havranek, Maximilian M},
  journal={BMC Medical Research Methodology},
  volume={23},
  number={1},
  pages={280},
  year={2023},
  publisher={BioMed Central}
}

@article{austin2024using,
  title={Using Multilevel Models and Generalized Estimating Equation Models to Account for Clustering in Neurology Clinical Research},
  author={Austin, Peter C},
  journal={Neurology},
  volume={103},
  number={8},
  pages={e209947},
  year={2024},
  publisher={AAN Enterprises}
}

@article{berta2024uncover,
  title={Uncover mortality patterns and hospital effects in COVID-19 heart failure patients: a novel Multilevel logistic cluster-weighted modeling approach},
  author={Berta, Paolo and Levantesi, Susanna and Marta, Gianmarco},
  journal={arXiv preprint arXiv:2405.11239},
  year={2024}
}

@article{frontiers2023machine,
  title={Machine learning-based prediction of hospital prolonged length of stay admission at emergency department: a Gradient Boosting algorithm analysis},
  author={Ciaroni, Silvia and Maggiotto, Giorgio and Moro, Jessica and Calza, Stefano and Martino, Gian Paolo and Benvenuto, Marco and De Santis, Marco and Scaglione, Francesco},
  journal={Frontiers in Artificial Intelligence},
  volume={6},
  pages={1179226},
  year={2023},
  publisher={Frontiers Media SA},
  doi={10.3389/frai.2023.1179226}
}

@article{bmcmedinform2021prediction,
  title={The prediction of hospital length of stay using unstructured data},
  author={Chrusciel, Jan and Girardon, Fran{\c{c}}ois and Roquette, Laure and Laplanche, David and Duclos, Antoine and Sanchez, St{\'e}phane},
  journal={BMC Medical Informatics and Decision Making},
  volume={21},
  number={1},
  pages={351},
  year={2021},
  publisher={BioMed Central},
  doi={10.1186/s12911-021-01722-4}
}

@article{young1973prediction,
  title={Prediction of hospital length of stay},
  author={Young, W John},
  journal={Health Services Research},
  volume={8},
  number={4},
  pages={287--295},
  year={1973},
  publisher={Health Research \& Educational Trust}
}

@article{sendak2020machine,
  title={Machine learning in medicine: a practical approach},
  author={Sendak, Mark P and Balu, Suresh and Schulman, Kevin A},
  journal={Gastroenterology},
  volume={158},
  number={2},
  pages={2070--2084},
  year={2020}
}

\end{document}